\documentclass[10pt,letterpaper]{article} 
\usepackage{multirow}
\usepackage{times}
\usepackage{epsfig}
\usepackage{graphicx}
\usepackage{amsmath}
\usepackage{amssymb}
\usepackage{textcomp}
\usepackage{mathtools}
\usepackage{adjustbox}
\usepackage{amssymb,amsmath}
\usepackage{hhline}
\usepackage{enumitem}
\usepackage{lipsum}
\usepackage[table,xcdraw]{xcolor}
\usepackage{pdflscape}
\usepackage{tabularx}
\usepackage{booktabs}
\usepackage{comment}
\usepackage{xcolor}
\makeatletter
\newcommand{\mathleft}{\@fleqntrue\@mathmargin0pt}
\newcommand{\mathcenter}{\@fleqnfalse}
\makeatother
\usepackage[pagebackref=true,breaklinks=true,letterpaper=true,colorlinks,bookmarks=false]{hyperref}
\usepackage[capitalize]{cleveref}
\crefname{section}{Sec.}{Secs.}
\Crefname{section}{Section}{Sections}
\Crefname{table}{Table}{Tables}
\crefname{table}{Tab.}{Tabs.}
\usepackage{enumitem}
\setlist[itemize]{left=0em,label=--}

\title{Multi Sentence Description of Complex Manipulation Action Videos}

\author{
  \begin{tabular}{c}
    Fatemeh Ziaeetabar$^{1,2}$, Reza Safabakhsh$^{2}$, Saeedeh Momtazi$^{2}$, \\
    Minija Tamosiunaite$^{1,3}$ and Florentin W\"org\"otter$^{1}$ \\
  \end{tabular} \\
  $^{1}$ III. Physics Institute, University of G\"ottingen, Germany, \\
  {\tt\small \{fziaeetabar, worgott\}@gwdg.de} \\
  $^{2}$ Faculty of Computer Engineering, Amirkabir University of Technology, Iran, \\
  {\tt\small \{safa, momtazi\}@aut.ac.ir} \\
  $^{3}$ Faculty of Informatics, Vytautas Magnus University, Lithuania, \\
  {\tt\small minija.tamosiunaite@vdu.lt}
}

\begin{document}

\maketitle
\thispagestyle{empty}


\begin{abstract}
   Automatic video description requires the generation of natural language statements about the actions, events, and objects in the video.    An important human trait, when we describe a video, is that we are able to do this with variable levels of detail. Different from this, existing approaches for automatic video descriptions are mostly focused on single sentence generation at a fixed level of detail. Instead, here we address video description of manipulation actions where different levels of detail are required for being able to convey information about the hierarchical structure of these actions relevant also for modern approaches of robot learning.
We propose one hybrid statistical and one end-to-end framework to address this problem. The hybrid method needs much less data for training, because it models statistically uncertainties within the video clips, while in the end-to-end method, which is more data-heavy, we are directly connecting the visual encoder to the language decoder without any intermediate (statistical) processing step.
Both frameworks use LSTM stacks to allow for different levels of description granularity and videos can be described by simple single-sentences or complex multiple-sentence descriptions. In addition, quantitative results demonstrate that these methods produce more realistic descriptions than other competing approaches.
\end{abstract}

\section{Introduction}
\begin{figure*}[!h]
    \centering
    \includegraphics[width=0.96\textwidth]{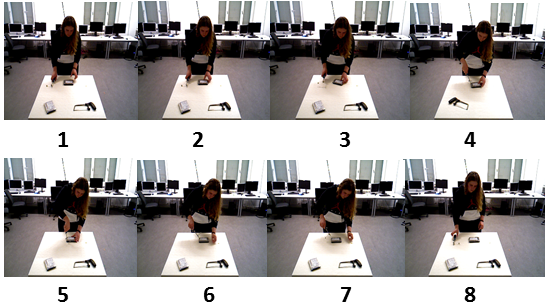}
    \caption{An example of the performance of the hybrid method in producing detailed, multiple- and single-sentence descriptions of a screwing scenario \cite{dreher2020learning} containing 8 video snippets (1,...,8). \textbf{Detailed sentences:}
\textit{For left hand:} (1) The left hand touches the top of a screwdriver on the table. (2) They untouch the table. (3) They move together above the table. (4) They touch a hard disk on the table. (5) They move together inside of a hard disk on the table. (6) They untouch the top of the hard disk and  move together above the table. (7) They touch the table. (8) The left hand untouches the top of the screwdriver on the table. \textit{For right hand:} (1, 2) Idle, (3) The right hand touches near (``around''-relation) the hard disk on the table. (4) The right hand continues to touch near the hard disk on the table. (5) The right hand untouches near the screwdriver from the table. (6, 7, 8) Idle. \textbf{Multiple sentences: } \textit{For left hand:} (1.2.3.4) The left hand picks up a screwdriver from the table and places it on a hard disk. (4.5) They perform screwing in the inside of the hard disk on the table. (6.7.8) The left hand leaves the screwdriver on the table. \textit{For right hand:} (3.4) The right hand keeps touching near the hard disk on the table. (5) The right hand leaves the hard disk on the table. \textbf{One sentence:} \textit{For left hand:} The left hand performs screwing inside of a hard disk on the table by a screwdriver. \textit{For right hand:} The right hand holds the hard disk and then lets go. }
    \label{snapshots}
\end{figure*}

\begin{figure}[!h]
    \centering
    \includegraphics[width=0.96\textwidth]{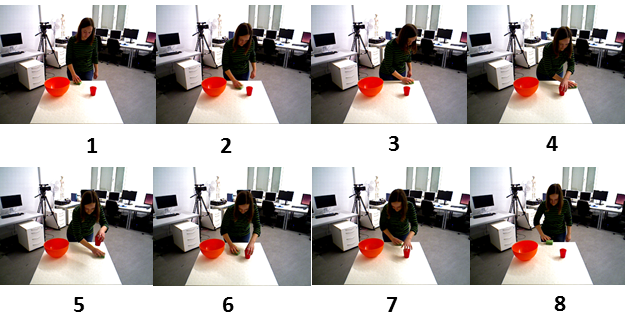}
    \caption{An example of the performance of our end to end method in producing detailed, multiple- and single-sentence descriptions of a wiping scenario \cite{dreher2020learning} containing 8 video snippets (1,...,8). \textbf{Detailed sentences:} \textit{For left hand:} (1) The left hand touches a sponge on a table. (2) They move together on the table. (3, 4, 5, 6, 7) They keep moving together on the table. (8) The left hand untouches the sponge on the table. \textit{For right hand:} (1, 2, 3) Idle.  (4) The right hand touches a cup on a table. (5) They untouch the table. (6) They move together above the table. (7) They touch the table. (8) The right hand untouches the cup on the table. \textbf{Multiple sentences:} \textit{For left hand:} (1) The left hand touches a sponge on a table. (2.3.4.5.6.7) It wipes the table. (8) The left hand leaves the sponge on the table. \textit{For right hand:} (4.5.6.7) The right hand picks and places a cup on the table. (8) The right hand then leaves the cup on the table. \textbf{One sentence:} \textit{For left hand:} The left hand wipes the table by a sponge. \textit{For right hand:} The right hand lifts and puts a cup on the table.
}
    \label{snapshots_end2end}
\end{figure}

Describing video content using natural language has emerged as a formidable challenge, capturing significant attention in recent years. Despite the considerable advancements showcased in the ``Related Works" section, the field continues to grapple with significant challenges, with deep learning-based methodologies predominantly driving progress.

This study specifically targets videos of human manipulation actions, which are complex but crucial for applications in robotics, especially for learning from demonstrations. Videos and language interactions between humans and robots are becoming more common, making it important for machines to understand and generate clear and relevant sentences. However, human manipulation actions are complex and need a careful approach to arrive at good video descriptions.

In response to this need, we have introduced two distinct methodologies to tackle this issue. The first is a hybrid statistical method, which adeptly combines video analysis techniques with the generation of simplified semantic representations (SRe), seamlessly integrated with a stack of Long Short-Term Memory (LSTM) networks. On the other hand, our second approach is an end-to-end trained LSTM-stack, designed to operate directly on the video data. While the latter demands a substantial volume of training data to function optimally, the hybrid method significantly reduces this requirement through its use of SRe-representations, without compromising on the quality of the generated descriptions. Both methods exhibit exemplary performance, surpassing existing state-of-the-art solutions in the field of video description generation.

Our approach stands out because it can create different levels of descriptions, from very detailed to more general statements about the video. This flexible approach, shown in Figs.~\ref{snapshots}, \ref{snapshots_end2end}, lets users choose the level of detail they need. Whether it is creating a detailed instruction manual from a video for training purposes or a general story for a movie, our method adapts to provide the right level of detail.

The capability to generate multi-level descriptions is of paramount importance in robotic applications and learning from demonstration. Robots often operate in diverse and dynamic environments, requiring them to understand and execute complex tasks. The multi-level representations (descriptions) generated by our method enable robots to comprehend the intricacies of human actions at various granularity levels, facilitating a more nuanced understanding and execution of tasks. Naturally, for any intrinsic understanding of actions and its sub-action the robot does not have to utter any sentence and only make use of the descriptive depth per se. This is especially beneficial in learning from demonstration, where robots learn to perform tasks by observing human actions. The multi-level representations provide a rich semantic context, aiding the robots in deciphering the intent and subtleties of human actions, leading to more accurate and efficient task execution. The possibility of uttering a respective phrase will then, however, also allow the machine to engage in discourse, possibly asking detailed questions about different action steps.


\section{Related Works}
\label{LR}
Research on video description has evolved into three main categories: ``classical methods," ``statistical algorithms," and ``deep learning-based approaches."

\paragraph{Classical Methods}

The earliest approaches to video description relied on SVO (Subject, Object, Verb) triple-based methods. These methods comprised two fundamental stages: content identification and text generation. Classical computer vision (CV) and natural language processing (NLP) techniques were employed to detect visual entities in videos and map them to standard sentences using handcrafted templates. However, these rule-based systems were effective only in highly constrained environments and for short video clips \cite{kojima2002natural, hanckmann2012automated}.

\paragraph{Statistical Algorithms}

To overcome the limitations of rule-based methods, researchers turned to statistical approaches. These methods utilized machine learning techniques to convert visual content into natural language descriptions using parallel corpora of videos and their associated annotations \cite{rohrbach2013translating, guadarrama2013youtube2text}. They initially involved object detection and action recognition through feature engineering and traditional classifiers, followed by the translation of retrieved information into natural language using Statistical Machine Translation (SMT) techniques \cite{koehn2007moses}. However, this separation of stages made it challenging to capture the interplay between visual features and linguistic patterns, limiting the ability to learn transferable relationships between visual artifacts and linguistic representations \cite{aafaq2019video}.

\paragraph{Deep Learning-Based Approaches}

The success of deep learning in both computer vision (CV) and natural language processing (NLP) inspired researchers to adopt deep learning techniques for video description tasks. Vinyals et al. \cite{vinyals2015show} introduced an image description model based on an encoder-decoder architecture, using Convolutional Neural Networks (CNNs) as image encoders and Long Short-Term Memory networks (LSTMs) as language decoders. Building on this, Venugopalan et al. \cite{venugopalan2014translating} proposed a similar framework for video annotation, where features were extracted from each video frame, and LSTM layers were used to generate textual descriptions. However, this approach struggled to capture the temporal order of events in a video.

Subsequent advancements included Pan et al.'s unified framework with visual semantic embedding \cite{pan2016jointly}, which extended to exploit temporal information in videos \cite{pan2016hierarchical}. Yu et al. \cite{yu2016video} introduced hierarchical Recurrent Neural Networks (RNNs) for video captioning. Wang et al. \cite{wang2018reconstruction} designed a reconstruction framework with a novel encoder-decoder architecture, leveraging both forward (video to sentence) and backward (sentence to video) flows for video captioning.

Krishna et al. \cite{krishna2017dense} suggested a two-stage method for describing short and long events, where events in a video were first detected, and descriptions were generated based on event dependencies. Nian et al. \cite{nian2017learning} presented an enhanced sequence-to-sequence model for video captioning, incorporating a mid-level video representation method known as the video response map, in addition to CNN features from sampled frames.

Recent advancements in video description have extended beyond traditional deep learning-based methods, starting the new era of transformer-based models. These transformer-based approaches have emerged as a distinct branch within the realm of deep learning, offering unique and promising capabilities for generating coherent and contextually relevant text descriptions for videos. 

\paragraph{Transformer-Based Paradigms}

Pioneering this paradigm shift, Ji et al. introduced the groundbreaking Action Genome framework \cite{ji2020action}. Within this framework, actions are represented as intricate compositions of spatio-temporal scene graphs. This innovative approach empowers the generation of rich and structured descriptions that encapsulate the nuances of human activities, providing a fresh perspective on video description.

Subsequently, Kim et al. presented ``ViLT" \cite{kim2021vilt}, a vision-and-language transformer model that redefines the landscape of video captioning. What sets ViLT apart is its ability to autonomously learn the art of generating video descriptions, all without relying on traditional convolutional layers or region supervision. This marks a significant leap towards harnessing the full potential of transformers in the domain of video captioning.

Another noteworthy contender in the transformer-based video captioning arena is ``SwinBERT" \cite{lin2022swinbert}. This model incorporates sparse attention mechanisms, a novel approach that enhances efficiency without compromising performance. SwinBERT aims to strike an optimal balance between computational efficiency and captioning accuracy, contributing to the ongoing evolution of video description methods.

Looking ahead, the field of transformer-based video description continues to evolve, with ongoing research exploring new directions and potential solutions. Recent developments include Fu et al.'s framework as a fully end-to-end VIdeO-LanguagE Transformer (VIOLET), which adopts a video transformer to explicitly model the temporal dynamics of video inputs \cite{fu2021violet}. Then then present an extension of the VIOLET framework \cite{fu2021violet} where the supervision from Masked visual modeling (MVM) training is backpropagated to the video pixel space \cite{fu2023empirical}. In another study, Gu et al. \cite{gu2023text}  proposes a two-stream transformer model, TextKG, for video captioning. One stream focuses on external knowledge integration from a pre-built knowledge graph to handle open-set words, while the other stream utilizes multi-modal information from videos. Cross-attention mechanisms enable information sharing between the streams, resulting in improved performance.

Transformer-based models excel at generating coherent single-sentence video descriptions but may face challenges when crafting multi-level descriptions of manipulation actions, especially in complex scenarios. Transformers are inherently optimized for single-sentence descriptions and may not naturally accommodate the intricacies of manipulation actions, which often involve decomposing actions into atomic components and organizing them into complex sequences.

By contrast, our proposed hierarchical approach is tailored to address the nuances of manipulation actions, offering the flexibility to generate multi-level descriptions that span from detailed atomic actions to more concise narratives.

While transformer-based methods are strong in language generation and comprehension, direct comparisons with our approach may not be straightforward due to their single-sentence focus. The choice between these approaches should align with the specific task requirements, whether it entails generating single-sentence descriptions or multi-level, detailed descriptions of manipulation actions.

Note that at the end of the Methods section we are discussing some more approaches in greater detail, which we are using for comparison with our results.

\begin{figure}[!h]
    \centering
    \includegraphics[width=0.96\textwidth]{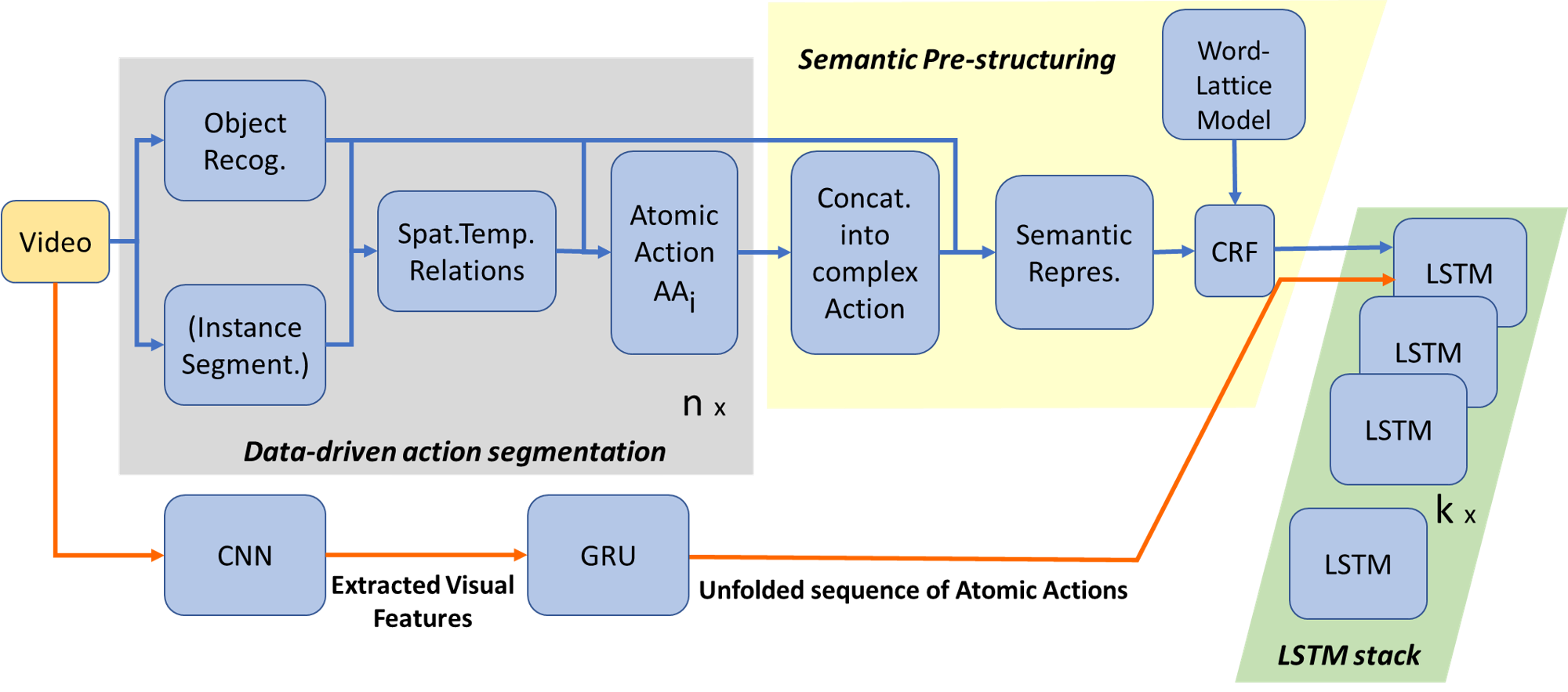}
    \caption{Flow diagram of both methods. Top (blue arrows): Hybrid method, bottom (orange arrows): end-to-end method, where the end
to end method uses the temporal (start- and end-) information from the segregation into complex actions from the top-path. Similar to \cite{venugopalan2015sequence} we unroll the LSTMs to a fixed 80 frames (using zero padding or frame drop in case of too short or too long video snippets) as this number offers a good trade-off between memory consumption and the ability to feed many video frames to the LSTM. The operations in the gray box are repeated n times, where n is the number of atomic actions needed to desribe one given complex action. Furthermore, we use a stack of k LSTMs to arrive at different levels of description granularity. The role of both variables is explained in more detail in the text. }
    \label{Flow}
\end{figure}

\section{Methods}
\label{overview}
The goal of the study is to provide a framework for describing manipulation actions in video using human language sentences at different levels of granularity. We will here compare two new methods. One of our methods (Method 1) first computes novel semantic representations capturing spatio-temporal relations between objects from the video stream and then uses them as a front-end to a neural network. The other (Method 2) is directly using an end-to-end-trained network combined with a novel approach for generating action proposals in time. Figure~\ref{Flow} provides an overview over these methods where we will in the following sections describe their different components in detail.


\begin{figure}[!b]
    \centering
    \includegraphics[width=0.97\textwidth]{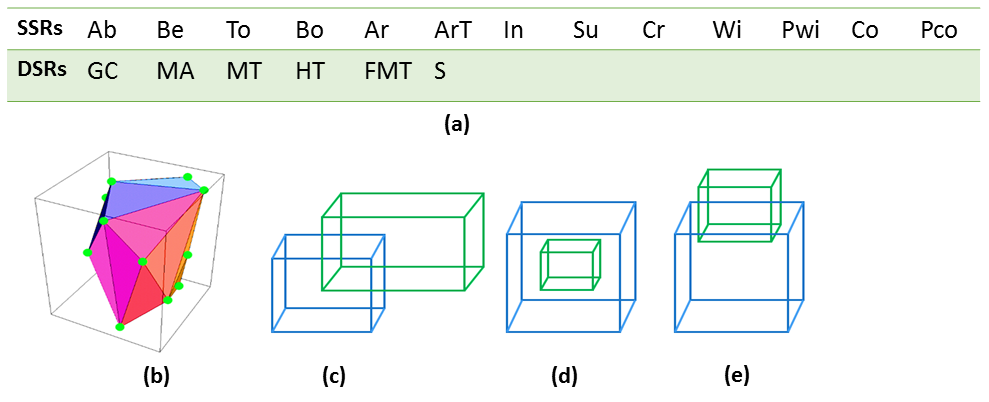}
    \caption{(a) The list of static spatial relations (SSRs) are ``Above (Ab)'', ``Below (Be)'', ``Around (Ar)'', ``Inside (In)'', ``Surround (Su)'', ``Cross (Cr)'', ``Within (Wi)'', ``Partial within (Pwi)'', ``Contain (Co)'' and ``Partial contain (Pco)'' while ``Above'', ``Below'' and ``Around'' relations in combination with ``Touching'' were converted to ``Top (To)'', ``Bottom (Bo)'' and ``Touching Around (ArT)'', respectively. The dynamic spatial relations (DSRs) are ``Getting close (Gc)'', ``Moving apart (Ma)'', ``Moving together (Mt)'', ``Halting together (Ht)'', ``Fixed-moving together (Fmt)'' and ``Stable (S)'', (b) A sample of a convex hull, (c-e) samples of some static spatial relations defined in this paper. If we assume the green cube as $ \alpha $ and the blue cube as $ \beta $, then: (c) $ SSR(\alpha,\beta)=Cr $ and $ SSR(\beta,\alpha)=Cr $, (d) $ SSR(\alpha,\beta)=Wi $ and $ SSR(\beta,\alpha)=Co $, (e) $ SSR(\alpha,\beta)=Pwi $ and $ SSR(\beta,\alpha)=Pco $ }
    \label{list_SR}
\end{figure}

\section{Method 1: Hybrid Approach}
This method embodies a hybrid model that synergistically integrates a statistical semantic pre-analysis of actions with the temporal processing capabilities of Long Short-Term Memory (LSTM) networks. The statistical approach aims at distilling the inherent semantics of various actions, facilitating a structured breakdown of complex actions into simpler, atomic units. Subsequent processing by the LSTM aids in capturing the temporal dependencies and nuances within sequences of actions. A significant advantage of this hybrid approach lies in its efficiency: it demands a relatively limited dataset for effective training, making it feasible for applications where large volumes of annotated data might be scarce or challenging to obtain.

\subsection{Object Characteristics and Pre-processing (Fig.~\ref{list_SR})} 

To craft accurate and descriptive sentences about a given environment, object recognition is pivotal. This happens in two distinct stages:

\begin{enumerate}
    \item \textbf{2D Pre-processing Using RGB Images}:
    \begin{itemize}
        \item Objects are detected employing YOLO \cite{redmon2018yolov3}, which has been trained on the labeled objects from the pertinent datasets.
        \item Given the intricate motion and positions of human hands, OpenPose \cite{cao2017realtime} is requisitioned for their detection. Utilizing key points delineated by OpenPose, a 2D bounding box for each hand is determined.
    \end{itemize}
The result of this stage is a compilation of 2D bounding boxes: one set representing objects identified by YOLO and another for hands discerned via OpenPose.
    
    \item \textbf{3D Processing and Spatial Relation Inference}:
    \begin{itemize}
        \item \textit{Processing with Depth Information}: For datasets that contain depth metrics, the 2D data harvested from the preceding stage is amalgamated with point clouds extracted from depth images. This synthesis leads to the generation of 3D bounding boxes for objects, facilitating a more granular comprehension of their spatial inter-relationships.
        
        \item \textit{Inferring 3D Relations from 2D Imagery}: 
For datasets that solely offer 2D information without accompanying depth data, our method taps into the potential of the 3D-R2N2 approach \cite{choy20163dr2n2}. The essence of the 3D-R2N2 method is its capacity to employ deep learning models to reconstruct 3D objects from 2D images. More specifically, it utilizes a recurrent neural network (RNN) that integrates information from multiple views of an object (or even a single view) to produce a consistent 3D shape. This network is trained using a combination of synthetic and real images, allowing it to make  predictions about the 3D structure of objects seen in diverse 2D images. Hence, even in the absence of explicit depth information, 3D spatial relationships between objects can be inferred. 
    \end{itemize}
\end{enumerate}
The result of this is a three-dimensional representation of the scene, which augments the system's spatial awareness and allows for the completion of atomic action quintuples, as a three-dimensional spatial relationship is a fundamental element of an atomic action's definition.

\label{atomic}
Our method relies on a purely data driven recognition of actions using computer vision. This is achieved by recognizing the time-chunks and the  structure of so-called \textit{atomic actions}, which are at a later stage concatenated into real action-names. Atomic actions are defined by the following quintuple: \textit{(Subject, Action-Primitive, Object, Spatial Relation, Place)}. For example, if a hand touches the top of a cup on a table, its corresponding quintuple will be: (hand, touch, cup, top, table). In the following, we will give a brief reference to each of these five items.

\begin{enumerate}
  \item \textbf{The subject} has two states and can be \textit{``Hand (H)''} or \textit{``Merged entity (Me)''}. Merged entity is used when the hand and its touched object act as a same entity and perform an action on another object.

  \item \textbf{The action primitive} has four states. Two of those are \textit{``Touching (T)''} and \textit{``Untouching (U)''}, which are used when one object starts touching or untouching another object, respectively. The next two states are \textit{``Moving together (Mt)''} and \textit{``Fixed moving together (Fmt)''} for the times that two objects move together or one is fixed and the other is moving on the fixed one. These primitive states are enough to create the required semantic representations, where real actions, like cutting, stirring, etc., need to be captured (see below).

  \item \textbf{The object} has  four constituents, which are ``$(O_{1,2,3}),~G$'' (Object 1,2,3 and Ground) where the latter supports all other objects except the hand in the scene.
  $O_1$, $O_2$ and $O_3$ are the objects which are the first, second and third to display a change in their T/N relations, respectively. 
In \cite{ziaeetabar2018recognition}, it has been discussed that there are never more than these four constituents (+ the hand) existing in any manipulation action.
  \item \textbf{The spatio-temporal relation} between two objects (e.g. subject and object) has thirteen states (see Fig. \ref{list_SR}). Below we describe how those are computed.

  \item \textbf{The place} has five states. An action can occur on the surface of another object ($O_1$ to $O_3$), on the ground, or in the air \textit{(Air)}.
\end{enumerate}

\label{spatial_reasoning}
\subsection{Spatio-temporal relations: Item 4 from above}
\label{convex_hull}
To perform  this step we determine the convex hulls of all objects recognized in the video using the ``gift wrapping algorithm \cite{sugihara1994robust}''. Then we compute their spatio-temporal relations by standard set-calculus (see Supplement).

Consequently a set of static as well as dynamic relations between the different objects can be determined in a straight forward manner summarized in Fig. ~\ref{list_SR}.

\subsection{Mapping complex actions to atomic actions}

Above we had described that we use action primitives with only four states as one component in the atomic action tuple, where these four states can all be recognized directly in a video. Real actions, to which one needs to associate a characteristic action word for video description, like cutting, stirring, pick\&place, consist of a string of atomic actions. In a pre-processing step we used a context free grammar (see Supplement) to decompose real actions into their constituting atomic actions. As a result any complex action, for example ``cut'' is then represented as a string of atomic actions $AA_i$, e.g., $cut=[AA_1, AA_2,\dots , AA_n]$. This processing step had been done offline and the resulting mappings have been stored in a so-called \textit{library of action mappings}. 


\subsection{Semantic Representations: SRe}
Semantic representations use the same 5-tuple structure as in the atomic action: \textit{(Subject, Action, Object, Spatial Relation, Place)}. Different from atomic actions, in the SRe real action(-names) are being used. This creates a nested system of the kind, f.e.
\begin{equation}
\begin{split}
SRe_{cut}=(Subject, cut=[AA_1,AA_2,\dots AA_n],\\ Object, Spatial Relation, Place)    
\end{split}    
\end{equation}
where each atomic action takes the structure of AA=(Subject, Action Primitive, Object, Spatial Relation, Place). Evidently, entities from atomic actions can reoccur in the SRe. Importantly, atomic actions can be easily recognized from video, because of the small number and simple structure of the action primitives (only 4). SRes are then derived from them as a string of AAs. As mentioned above we used a context free grammar (see Supplement) to pre-compute and decompose all complex actions in the data set into their constituting atomic action.

\subsection{The core of the hybrid statistical method}
\label{hybrid_s}
As a main advantage, the hybrid method works with a modest amount of data. In the hybrid method, first we generate the semantic representation (SRe) of the visual content including subject (tool), action, object, spatio-temporal relations and place, then we model the semantic relationships between the visual components through learning a Conditional Random Field (CRF) structure. Finally, we propose to formulate the generation of natural language description as a decoding stage using two layers of LSTMs. The LSTM framework allows us to model the video as a variable length input stream and creates output as a variable length sentence.

The start of a  manipulation actions can be defined when the hand touches something and the end when it is free again \cite{ziaeetabar2018recognition}.  This way we can break a long video into snippets that contain only one action. Then we associate every component of the SRe-5-tuple to a graph node and create for every video snipped a fully connected graph with 5 nodes. The edges between the nodes will after training contain the probabilities of a co-occurrence of two nodes. For example in a cutting-SRe it shall be more likely to find ``Subject=hand+knife'' (i.e. the subject is here a merged entity) together with ``object=apple'' than with ``object=lamp''. 

The training of the Conditional Random Field (CRF) stands is central to this method, marked by the need to choose an appropriate dataset. We draw upon 'The KIT Bimanual Actions Dataset' (\cite{dreher2020learning}) from which the CRF learns to extract relationships and to determine probabilities for all existing Semantic Representations (SRes). We emphasize that our choice of dataset is a calculated one, focused on the deployment of a quite limited dataset. This constraint serves to imbue our model with precision, preventing it from lapsing into an over-generalized state.

However, the decision to employ a limited dataset, while advantageous in terms of focus, introduces a potential challenge. With the restricted data at hand, there exists the risk of overestimating the probabilities of connections between nodes within our graphs. Such an overestimation can lead to inaccuracies when the CRF tries to predict connections for novel, previously unseen data. To mitigate this risk, we employ the Word Lattice Model (\cite{dyer2008generalizing}).

Formally, the Word Lattice Model is presented as a Directed Acyclic Graph, encapsulating an array of potential outcomes, each accompanied by a corresponding confidence level. Its primary utility, traditionally  in the domain of machine translation, extends beyond decoding the most probable hypothesis. This is, because it considers alternative interpretations that, while bearing slightly lower confidences, remain plausible. Within our methodology, the Word Lattice Model emerges as a secondary line of verification. It scrutinizes the probabilities elucidated by the CRF and juxtaposes them with its own cartography of potential relationships. Should the probabilities derived from the CRF starkly contradict those inferred by the Word Lattice Model, we invoke a re-adjustment process. This entails a reordering of probabilities to ensure their alignment with empirical and logical expectations.

The steps of the re-adjustment process are as follows:

\begin{enumerate}
    \item \textbf{Discrepancy Evaluation:} Initially, we perform a quantitative analysis to identify any significant discrepancies between the CRF's predicted probabilities and those of the Word Lattice Model. This analysis includes statistical tests to ascertain whether the differences are beyond acceptable margins, thereby warranting adjustments.
  \item \textbf{Identification of Outliers:} Probabilities that exhibit substantial deviations from those proposed by the Word Lattice Model are flagged as outliers. These outliers are then thoroughly examined to understand the underlying causes, such as data sparsity or bias in the training set.
\item \textbf{Probability Recalibration:} The recalibration involves adjusting the outlier probabilities. The steps involved in this recalibration are as follows:
\begin{itemize}
        \item \textbf{Empirical Evidence Gathering:} We revisit the training set to gather instances that correspond to the outlier probabilities. This involves analyzing segments where the co-occurrence of nodes is either over-represented or under-represented.
  \item \textbf{Reassessment of Contextual Factors:} Each instance is evaluated in its contextual entirety, considering factors such as the frequency of action-object pairings in various contexts and the diversity of the actions performed.
 \item \textbf{Expert Verification:} Subject matter experts review the instances to confirm or dispute the validity of the co-occurrence probabilities. Their insights are crucial for determining whether to retain, increase, or decrease specific probability values.
 \item \textbf{Adjustment Based on Consensus:} If a consensus is reached that the current probability values are inconsistent with the empirical evidence and expert opinion, we employ a mathematical model to calculate the new probability. This model incorporates the frequency of the observed co-occurrences and the experts' qualitative assessments to produce a more balanced and representative probability value.
\item \textbf{Integration and Normalization:} The newly calculated probabilities are integrated back into the CRF's model, followed by a normalization process to maintain the probabilistic model's consistency.
\end{itemize}
These steps ensure that each adjustment is not an arbitrary decision but is instead grounded in actual data and expert validation, leading to a CRF model that is both accurate and reliable.
 \item \textbf{Validation and Iteration:} Following recalibration, we conduct a validation phase using a held-out validation dataset, which serves as a new reference point for assessing the accuracy of the adjusted probabilities. If the validation results indicate that the adjustments have not yielded an improvement, we iteratively refine the probabilities. This iterative process is guided by a combination of empirical evidence and the feedback provided by the validation phase until the discrepancies are resolved.
 \item \textbf{Integration and Finalization:} Once the adjusted probabilities pass the validation phase, they are integrated back into the CRF model. The final model is then subjected to a comprehensive evaluation to ensure that the adjustments have enhanced its predictive performance and generalization capabilities.
\end{enumerate}
Through this rigorous re-adjustment process, we maintain the integrity and applicability of our CRF model, ensuring that it accurately reflects the complex relationships within the data, despite the limitations imposed by a smaller training set.

For readers seeking a more profound understanding of the nuanced mechanics of the Word Lattice Model, we refer to \cite{dyer2008generalizing}, which provides a comprehensive exploration of its intricacies.

Following this data preparation process, where we carefully extracted, verified, and fine-tuned our training samples, these curated datasets are now incorporated into the initial layer of the Long Short-Term Memory (LSTM) decoder. This step serves as the gateway to a complex series of computations, laying the foundation for the generation of detailed and descriptive narratives.

While most existing video description methods produce only one sentence for each video clip, an important capability that we  provide with our approach is the possibility of generating descriptions in the form of several coherent sentences for each action with different levels of descriptive detail. For this purpose, we use $k$ (Fig.~\ref{Flow}, right) parallel levels of LSTMs, where each LSTM level has two layers. To capture the lowest, most detailed descriptive level, the total number of LSTMs needed should equal the most complex string of atomic actions used to describe any real action in the data sets. The pre-analysis performed by us to create the \textit{library of action mappings} yielded that the most complex action consisted here of 14 AAs. Hence, 14 levels of LSTMs are maximally needed. At the first, bottom level descriptions consist of a set of very detailed sentences, each of which describes just one single atomic action. Obviously such sentences are non-human like and very awkward, but we can now begin to concatenate atomic actions, where this process is guided by the entries in the \textit{library of action mappings}, leading to sentences that describe concatenated groups of AAs by single action verbs. This is done until full concatenation. For every level that results this way one LSTM is trained, where often levels are left out due to the fact that not all AAs can be concatenated into a meaningful verb. As a result, this algorithm can produce coherent sentences describing manipulation videos at different levels.

\section{Method 2: End-to-End Network}
\label{e2e}

To address these constraints, we propose an end-to-end trainable framework designed to learn not only the temporal structure of input video sequences but also the intricate sequence modeling required for generating comprehensive multi-level descriptions. In the following sections, we will provide a detailed breakdown of the key components and steps in our end-to-end framework.

\subsection{Data Collection and Annotation}

Our atomic action recognition model relies on a selected collection dataset of videos which spans diverse domains to ensure a rich variety of actions, enhancing the model's versatility.  For this we carefully selected a set of five diverse datasets: YouCook2, MSR-VTT, ActivityNet Captions, EPIC-Kitchens, and the KIT Bimanual Action Dataset, to ensure the richness and versatility of our training data. These datasets cover a wide range of activities and domains, providing a comprehensive representation of human manipulation actions. This diversity is essential for enhancing our atomic action recognition model's adaptability and robustness. To harness the full potential of these datasets and create a reliable foundation for our model, we employed a meticulous data annotation process. This process involves human annotators who systematically analyzed each video in these datasets. They followed specific guidelines and the atomic action definitions to label and describe actions within the videos. This annotation process is essential to provide precise information about the sequence of atomic actions in each video forming the basis for our atomic action recognition model's training.

    

\subsection{Automatic Atomic Action Recognition}

In this section, we present an overview of our automatic end-to-end atomic action recognition and description approach, which consists of three integral stages:

\begin{enumerate}
\setcounter{enumi}{0}
    \item \textbf{Visual Feature Extraction (CNN):}
\end{enumerate}

To initiate the process, we employ Convolutional Neural Networks (CNNs) for visual feature extraction. This stage serves as the foundation for our atomic action recognition and description approach. The choice of CNNs is rooted in their proven effectiveness in handling spatial information, a fundamental requirement for accurate atomic action recognition. Atomic actions, as elemental components of video narratives, often exhibit fine-grained spatial intricacies, such as subtle hand movements, object interactions, or body postures. CNNs are renowned for their ability to capture intricate spatial patterns, making them a natural and robust fit for our problem.

The input to this stage is a set of video frames \( F = \{f_1, f_2, \ldots, f_{n_f}\} \), where \( n_f \) represents the number of frames in the video sequence. These frames contain rich visual information, encompassing the dynamic interplay of objects, scenes, and actors in the video. Our objective is to extract meaningful visual features from these frames.

To achieve this, we leverage the Visual Geometry Group (VGG) architecture, specifically exploring variants such as VGG16 and VGG19. VGG architectures are well-regarded for their architectural simplicity and remarkable proficiency in capturing spatial details. They have demonstrated exceptional performance in image classification tasks, making them a compelling choice for feature extraction in our context.

The output of this stage is a sequence of feature vectors \( X = \{x_1, x_2, \ldots, x_{n_f}\} \), where \( x_i \) represents the feature vector extracted from frame \( f_i \). These feature vectors encode the spatial characteristics and patterns observed in the video frames. They encapsulate information related to the distribution of edges, textures, shapes, and object arrangements within each frame.


\begin{enumerate}
\setcounter{enumi}{1}
    \item \textbf{Temporal Modeling (GRUs):}
\end{enumerate}

In this stage, we delve into the modeling of nuanced temporal dynamics within atomic actions using Gated Recurrent Units (GRUs). This is a pivotal step, as atomic actions often involve subtle and temporally precise movements that necessitate sophisticated modeling.

The input to this stage is the sequence of feature vectors \( X = \{x_1, x_2, \ldots, x_{n_f}\} \), which were extracted by the CNNs during the ``Visual Feature Extraction (CNN)" stage. Each feature vector \( x_i \) encodes the spatial characteristics of the corresponding frame \( f_i \).

Our objective is to model the temporal evolution of these feature vectors, capturing how atomic actions unfold over time. To achieve this, we employ GRUs, which are recurrent neural networks well-suited for handling sequential data.

The key operations of GRUs can be described as follows:

For each frame \( f_i \), the GRU computes the hidden state \( h_i \) using the input feature vector \( x_i \) and the previous hidden state \( h_{i-1} \). This operation is expressed as:

\[
h_i = \text{GRU}(x_i, h_{i-1})
\]

Here, \( h_i \) represents the hidden state at time step \( i \), and \( h_{i-1} \) is the hidden state from the previous time step. This recurrent operation enables the network to capture the temporal dependencies within the sequence of feature vectors.

The output of this stage is a sequence \( Y = \{y_1, y_2, \ldots, y_{n_f}\} \), where \( y_i \) represents the output at time step \( i \). These outputs \( y_i \) encapsulate the temporal dynamics of the input feature vectors. Each \( y_i \) can be thought of as a rich representation that encodes how the features within atomic actions evolve over time.

In essence, the ``Temporal Modeling (GRUs)" stage transforms spatial information into a temporal representation, allowing us to capture the temporal dynamics inherent in a sequence of atomic actions. The output sequence \( Y \) serves as a valuable foundation for the subsequent stages of our end-to-end approach, enabling us to generate comprehensive video descriptions.

\begin{enumerate}
\setcounter{enumi}{2}
    \item \textbf{Description Generation (Stacked LSTMs):}
\end{enumerate}

With the temporal modeling stage completed, we now focus on generating descriptive annotations for the video frames. This phase bridges the gap between visual data and textual descriptions.

The input to this stage is the sequence of output vectors \( Y = \{y_1, y_2, \ldots, y_{n_f}\} \) from the GRUs. However, instead of generating a single sentence, we employ a hierarchical approach to create multi-level descriptions.

Our hierarchical description generation system consists of \(k\) levels of Long Short-Term Memory (LSTM) layers, each with two layers as mentioned. This hierarchy enables us to produce descriptions at various levels of detail. 



Each level of the hierarchy corresponds to a specific level of detail in the descriptions. We train one LSTM for each level, often leaving out levels that do not result in meaningful concatenation due to the nature of the atomic actions.

Formally, the distribution over the output sequence \( (W) \) given the input sequence \( (F) \) can still be defined as \( P(W|F) \), where:

\[
p(w_1, w_2, \ldots, w_{n_w} | f_1, f_2, \ldots, f_{n_f}) = \prod_{t=1}^{n_w} p(w_t | w_{t-1}, h_t)
\]

Here, \( p(w_t | w_{t-1}, h_t) \) represents the distribution over the words in the vocabulary for predicting the next word \( w_t \) given the previous word \( w_{t-1} \) and the corresponding hidden state \( h_t \) of the LSTM at the current hierarchical level.

During training in the decoding stage, our objective is to estimate the model parameters to maximize the likelihood of the predicted output sentence given the hidden representation of the visual frame sequence and the previous words it has processed.

This hierarchical approach allows us to provide multi-level descriptions, ranging from detailed atomic actions to more concise and contextually organized narratives, enhancing the expressiveness and adaptability of our description generation system.

In summary, our end-to-end approach integrates VGG-based CNNs for robust visual feature extraction, GRUs for modeling intricate temporal dependencies, and stacked LSTMs for generating descriptive annotations. This multi-stage process enhances the precision and efficiency of our automatic annotation generation system, enabling it to identify subtle, temporally precise, and contextually significant atomic actions within video sequences.

Figure~\ref{Flow} depicts our end-to-end model at the bottom, as well as shows its relation to the hybrid approach.

\section{Evaluation Metrics and Methods for Comparison}

\subsection{Metrics}
Evaluation of video descriptions is a challenging task as there is no specific ground truth or ``right answer'', that can be taken as a reference for benchmarking accuracy. A video can be correctly described in a wide variety of sentences, that may differ not only syntactically but also in terms of semantic content \cite{aafaq2019video}. In this work, we report the results from two points of view: \textbf{(1)} applying automatic measures which have a high correlation with human judgement, such as ``Bilingual Evaluation Understudy ($BLEU$) \cite{papineni2002bleu}'', while BLEU@N (N=1 to 4) metric is used to calculate the matched N-grams between machine-generated and reference sentences, ``Metric for Evaluation of Translation with Explicit Ordering ($METEOR$) \cite{banerjee2005meteor}'' and ``Consensus based Image Description Evaluation ($CIDEr$) \cite{vedantam2015cider}'', \textbf{(2)} using human evaluation which is divided into three measurements: \textbf{(I)} grammatical correctness, \textbf{(II)} semantic correctness and \textbf{(III)} the relevance of the produced video description compared to what is present in the video.

These metrics are widely acknowledged for assessing the quality of generated textual descriptions in terms of their linguistic accuracy, relevance, and overall fluency.
\subsection{State of the Art Methods used for Comparison and Data Sets}

Here, we introduce several cutting-edge video annotation methods, with a specific focus on those designed for annotating manipulation actions. These methods will serve as key points of reference for a comparative analysis with our proposed approaches. We categorize these comparisons based on different aspects to provide a comprehensive evaluation of our method. Note, that in the Results section we have analysed the different methods in comparison to our approaches using the here (below) mentioned data sets. In addition to this we have analysed all methods using the KIT Bimanual Action Data Set.

\subsubsection{Multi-level Descriptions}

\textbf{\cite{ji2020action}} presents an innovative framework, ``Action Genome," for video understanding and description. This method focuses on actions within videos and delves into the intricate details of actions and their representations. It represents actions as compositions of spatio-temporal scene graphs, allowing for rich and structured descriptions of human activities. Our approach shares a common emphasis with ``Action Genome" on capturing nuanced actions within videos, albeit with different methodologies. While ``Action Genome" emphasizes scene graph-based representations, our approach employs a hierarchical system for generating multi-level descriptions. This allows us to capture fine-grained details within manipulation actions.

\textbf{\cite{krause2017hierarchical}} presents a hierarchical approach for generating descriptive image paragraphs. This method allows for multi-level descriptions, which aligns with our goal of generating multi-level descriptions for manipulation actions. Similar to our approach, it focuses on capturing fine-grained details within images and organizing them into coherent narratives.

\paragraph{Dataset:} YouCookII \cite{zhou2018automatic} contains cooking videos and is suitable for evaluating multi-level descriptions of manipulation actions. It allows us to assess how well your method can generate descriptions that capture actions at different levels of granularity within cooking tasks.

\subsubsection{LSTM-Based Encoding and Decoding}

\textbf{\cite{nabati2020video}} proposes a framework that includes two LSTM layers for encoding and decoding visual features and enhancing generated descriptions. This architecture bears resemblance to our approach, which also utilizes LSTM layers for both encoding and decoding stages. Additionally, both methods employ a module for word selection, with similarities to our word lattice module. This comparison highlights the commonality in utilizing LSTM-based encoding and decoding mechanisms.

\paragraph{Dataset:} MSR-VTT \cite{xu2016msr} is a diverse dataset that includes video descriptions. It can be used to evaluate LSTM-based encoding and decoding methods for video annotation. .

\subsubsection{Semantic Summarization}

\textbf{\cite{singh2021efficient}} introduces an approach for summarizing visual information using a boundary-based key frames selection method. This aligns with our semantic approach, where we summarize visual information using semantic representations (SRes) before translating them into human language sentences. Both methods focus on the effective summarization of visual content, albeit through different techniques.

\paragraph{Dataset:} ActivityNet Captions \cite{krishna2017dense} provides video segments with descriptions and can be used for evaluating semantic summarization. This dataset allows you to assess how well your method summarizes visual information into semantic representations and translates them into human language sentences.

\subsubsection{Robotic Manipulation Actions}

\textbf{\cite{nguyen2018translating}} is one of the few video description papers that specifically focus on manipulation actions. It translates videos into commands applicable for robotic manipulation using deep recurrent neural networks. Similar to our work, it concentrates on the description of manipulation actions, making it a relevant comparison. This emphasizes the shared focus on annotating videos containing manipulation actions.

\paragraph{Dataset:} EPIC-Kitchens \cite{damen2020epic} includes kitchen-focused videos and is suitable for evaluating methods related to robotic manipulation actions. It captures real-world cooking and manipulation actions in a kitchen environment, making it a good choice for assessing the annotation of such actions.

\subsubsection{All Methods checked with one Additional Dataset}

In tandem with the previously mentioned datasets, we have incorporated the \textbf{KIT Bimanual Action Dataset} \cite{dreher2020learning} as a crucial 3D dataset in our study. This inclusion serves to enhance the comprehensiveness of our research, with a specific focus on manipulation actions. All methods mentioned above have been analysed using this data set.


\section{Results and Discussion}
\label{results}
This section presents a comparative analysis of our proposed methods, the hybrid statistical method and the end-to-end method, against state-of-the-art approaches on related datasets.

\subsection{Multi-level Descriptions}
The evaluation of multi-level description generation capabilities was conducted on both, the curated data \cite{zhou2018automatic} and the full set of manipulation actions of YouCookII.

The subset comprised approximately 20\% of the total dataset, specifically chosen to include a diverse range of cooking activities and ingredient complexities, providing a representative sample of common kitchen tasks. This subset selection aims to simulate real-world scenarios where data may be limited but varied.

We juxtaposed the performance of our Hybrid Statistical Method (Method 1) and our End-to-End Method (Method 2) against the methods detailed by Ji et al. \cite{ji2020action} and Krause et al. \cite{krause2017hierarchical}, employing BLEU, METEOR, and CIDEr as our evaluation metrics. 

\begin{table}[ht]
\centering
\caption{Comparison of multi-level description generation performance using BLEU, METEOR, and CIDEr metrics on both a representative subset and the full set of manipulation actions in the YouCookII dataset \cite{zhou2018automatic}.}
\label{tab:multi_level_results}
\begin{tabularx}{\textwidth}{|X|c|c|c|c|c|c|}
\hline
\multirow{2}{*}{Method} & \multicolumn{3}{c|}{Subset dataset} & \multicolumn{3}{c|}{Full YouCookII dataset} \\ \cline{2-7} 
                        & BLEU-4  & METEOR & CIDEr & BLEU-4  & METEOR & CIDEr \\
\hline
Ji et al. \cite{ji2020action}         & 0.32  & 0.23   & 0.41  & 0.37  & 0.28   & 0.45  \\
Krause et al. \cite{krause2017hierarchical} & 0.28  & 0.20   & 0.38  & 0.35  & 0.26   & 0.43  \\
Our Method 1 (Hybrid)   & \textbf{0.44}  & \textbf{0.29}   & \textbf{0.48}  & 0.40  & 0.23   & 0.49  \\
Our Method 2 (End-to-End) & 0.34  & 0.25   & 0.45  & \textbf{0.43}  & \textbf{0.32}   & \textbf{0.55}  \\
\hline
\end{tabularx}
\end{table}

The results presented in Table \ref{tab:multi_level_results} confirm the strengths of our Hybrid Statistical Method (Method 1) in scenarios with constrained data, underscoring its efficacy where data may be limited. This mirrors the few-shot learning achievements demonstrated by Ji et al. \cite{ji2020action}, where actions are decomposed into spatio-temporal scene graphs to encode activities hierarchically. However, our method refines this approach by capturing action-object interactions with greater granularity, significantly improving recognition accuracy in limited-data contexts.

By contrast, our End-to-End Method (Method 2) demonstrates its superior scalability with the expanded data set of the full subset of manipulation action in YouCookII dataset, achieving higher scores across all metrics. While Ji et al. \cite{ji2020action} lay the groundwork for hierarchical representation, our end-to-end approach capitalizes on this by incorporating a comprehensive understanding of atomic actions and temporal dynamics, which is crucial for the nuanced generation of video descriptions.

Furthermore, Krause et al. \cite{krause2017hierarchical} explore narrative generation through dense captioning of image regions, which our approach advances by translating into the video domain, thus managing to weave a coherent story over the sequence of frames. By doing so, our method maintains the intricate details and the broader context of the visual narrative, a challenge not fully addressed by the static image focus of Krause et al.
The seamless integration of visual perception with linguistic output in our End-to-End Method (Method 2) distinguishes it from the multi-stage processes typical of previous works. This streamlined pipeline preserves semantic integrity and ensures continuity in the narrative flow, resulting in video captions that are not only contextually accurate but also rich in detail.

\subsection{LSTM-Based Encoding and Decoding}
To evaluate the LSTM-based encoding and decoding strategies for video captioning, we focused on a manipulation-specific subset of the MSR-VTT dataset \cite{xu2016msr}. This subset was carefully curated to include a variety of interaction-intensive videos, offering a rigorous testing ground for our methods. Further, we extracted a 20\% segment of this manipulation-focused subset to assess the performance of our Hybrid Statistical Method (Method 1) under the constraints of limited data availability. This smaller sample reflects real-world scenarios where comprehensive datasets may not be readily accessible. In these conditions, our Hybrid Statistical Method demonstrated a notable advantage over our End-to-End Method (Method 2), underscoring its suitability for contexts with scarce data resources (see Table~\ref{tab:lstm_results}).

Our comparative analysis also included the Boosted and Parallel LSTM (BP-LSTM) architecture proposed by Nabati et al. \cite{nabati2020video}. While their architecture innovatively employs a boosted and parallel approach to enhance the LSTM's video captioning capabilities, our End-to-End Method leverages a more integrated processing pipeline. This cohesive approach facilitates a direct flow of information from the video content to the generated captions, optimizing the use of semantic cues and temporal dynamics. This mechanism allows for a nuanced discernment of relevant features across video frames, leading to captions that are not only syntactically and semantically coherent but also contextually richer and more descriptive. The results on the full manipulation subset of the MSR-VTT dataset indicate that our End-to-End Method outperforms the BP-LSTM, especially in generating accurate and detailed descriptions for complex manipulation actions.

\begin{table}[ht]
\centering
\caption{Comparison of LSTM-based encoding and decoding methods on both a representative subset and the full set of manipulation actions in the MSR-VTT dataset \cite{xu2016msr}.}
\label{tab:lstm_results}
\begin{tabularx}{\textwidth}{|X|c|c|c|c|c|c|}
\hline
\multirow{2}{*}{Method} & \multicolumn{3}{c|}{Subset dataset} & \multicolumn{3}{c|}{Full MSR-VTT dataset} \\ \cline{2-7} 
                        & BLEU-4 & METEOR & CIDEr & BLEU-4 & METEOR & CIDEr \\
\hline
Nabati et al. \cite{nabati2020video} & 0.22 & 0.16 & 0.25 & 0.28 & 0.19 & 0.44 \\
Our Method 1 (Hybrid)   & \textbf{0.35} & \textbf{0.22} & \textbf{0.55} & 0.33 & 0.21 & 0.43 \\
Our Method 2 (End-to-End) & 0.33 & 0.21 & 0.50 & \textbf{0.38} & \textbf{0.25} & \textbf{0.68} \\
\hline
\end{tabularx}
\end{table}

The comparative evaluation presented in Table \ref{tab:lstm_results} illuminates the strengths of our approach in relation to the BP-LSTM architecture proposed by Nabati et al. \cite{nabati2020video}. Their architecture introduces an ensemble of LSTM layers augmented with a boosting algorithm, representing a notable development in video captioning. Their method employs multiple LSTMs in a parallel configuration to encode and decode video data, aiming to enhance the precision of the generated textual descriptions.

In our work, we adopt a different perspective by simplifying the captioning process with an End-to-End Method. This approach may appear modest in comparison to the parallel and boosted networks utilized by Nabati et al., but it aligns closely with the inherent sequential processing strengths of LSTM networks. By refining the encoding scheme to better capture the temporal flow of video content, we reduce redundancy and focus the LSTM on the critical task of generating contextually rich captions.

Our architecture's strength lies in its ability to distill and utilize salient video features effectively, thereby enhancing the LSTM's capacity for predictive accuracy in caption generation. It is particularly adept at handling the intricacies of manipulation actions, which are essential for producing coherent and detailed narratives that resonate with the unfolding events in the videos.

When applied to the expansive dataset of MSR-VTT, our method consistently performs well, demonstrating its capability to scale and maintain performance without the need for complex, multi-stage frameworks.

\subsection{Semantic Summarization}
Semantic summarization of video content is pivotal for generating concise yet comprehensive descriptions. Our evaluation extends to the ActivityNet Captions dataset \cite{krishna2017dense}, focusing on a subset of manipulation actions and approximately 20\% of this subset to reflect limited-data scenarios.

Our comparison encompassed our Hybrid Statistical Method (Method 1) and our End-to-End Method (Method 2) against the boundary-based keyframes selection approach by Singh et al. \cite{singh2021efficient}. Their approach efficiently reduces computational costs by encoding visual information through selected keyframes, but this can lead to overlooking the temporal relationships essential for fully understanding the sequence of events within a video.

By contrast, our End-to-End Method maintains continuity of the video narrative, capturing the essence of actions and contexts without the information loss that may accompany keyframe reduction. Our Hybrid Statistical Method, tailored for sparse data environments, generalizes from limited examples to yield accurate summaries.

The results, presented in Table \ref{tab:semantic_summarization_results}, demonstrate that our methods surpass the keyframe-based summarization, especially in scenarios where maintaining narrative flow and temporal dynamics is critical for generating precise video descriptions. This is due to the comprehensive analysis of video content that our methods undertake, ensuring no vital information is missed during the summarization process.

\begin{table}[ht]
\centering
\caption{Comparison of semantic summarization performance on the ActivityNet Captions dataset.}
\label{tab:semantic_summarization_results}
\resizebox{\textwidth}{!}{
\begin{tabular}{|l|c|c|c|c|c|c|}
\hline
\multirow{2}{*}{Method} & \multicolumn{3}{c|}{Subset dataset} & \multicolumn{3}{c|}{ActivityNet Captions dataset} \\ \cline{2-7} 
                        & BLEU-4 & METEOR & CIDEr & BLEU-4 & METEOR & CIDEr \\
\hline
Singh et al.            & 0.25   & 0.20   & 0.45  & 0.30   & 0.23   & 0.50   \\
Method 1 (Hybrid)       & \textbf{0.32}   & \textbf{0.25}   & \textbf{0.60}  & 0.28   & 0.24   & 0.52   \\
Method 2 (End-to-End)   & 0.29   & 0.22   & 0.55  & \textbf{0.35}   & \textbf{0.27}   & \textbf{0.65}   \\
\hline
\end{tabular}
}
\end{table}

\subsection{Robotic Manipulation Actions}

Robotic manipulation commands derived from video data require a nuanced understanding of sequential actions and context. Our analysis leveraged the EPIC-Kitchens dataset \cite{damen2020epic}, scrutinizing a comprehensive set of manipulation actions and a representative 20

Nguyen et al. \cite{nguyen2018translating} introduced a method employing RNNs to translate video sequences into robotic commands. While their method capitalizes on advanced feature extraction and a two-layer RNN for improved accuracy, it may not fully capture the subtleties of manipulation actions within a kitchen environment.

Our Hybrid Statistical Method (Method 1) excels in scenarios with sparse data, effectively capturing relevant features and translating them into accurate commands. Conversely, our End-to-End Method (Method 2) demonstrates its superiority in a data-abundant environment by seamlessly integrating video frame analysis and command generation, thus preserving the contextual flow of actions.

The results, as summarized in Table~\ref{tab:robotic_manipulation_results}, suggest that our End-to-End Method outperforms Nguyen et al.'s approach when ample data is available, due to its comprehensive processing of sequential video frames, which is crucial for generating contextually rich commands. The Hybrid Statistical Method, on the other hand, showcases its robustness and adaptability in limited-data scenarios.

\begin{table}[ht]
\centering
\caption{Comparative analysis of translation methods for robotic manipulation actions using the EPIC-Kitchens dataset.}
\label{tab:robotic_manipulation_results}
\begin{tabularx}{\textwidth}{|X|c|c|c|c|c|c|}
\hline
\multirow{2}{*}{Method} & \multicolumn{3}{c|}{Subset dataset} & \multicolumn{3}{c|}{Full EPIC-Kitchens dataset} \\ \cline{2-7} 
                        & BLEU-4 & METEOR & CIDEr & BLEU-4 & METEOR & CIDEr \\
\hline
Nguyen et al.           & 0.28 & 0.19 & 0.75 & 0.26 & 0.20 & 0.50 \\
\hline
Our Method 1 (Hybrid)   & \textbf{0.34} & \textbf{0.25} & \textbf{0.65} & 0.33 & 0.21 & 0.57 \\
\hline
Our Method 2 (End-to-End) & 0.30 & 0.22 & 0.59 & \textbf{0.39} & \textbf{0.32} & \textbf{0.72} \\
\hline
\end{tabularx}
\end{table}

In summary, our End-to-End Method's integrated approach to processing and translating video data into manipulation commands accounts for its high performance on the full dataset. It manages to maintain the integrity and continuity of the video content, translating it into commands that are not just accurate in terms of individual actions, but also in their temporal and contextual relevance. This capability is particularly beneficial for robotic manipulation tasks that require an understanding of the sequence and context of actions. Our Hybrid Statistical Method's performance in the limited-data subset further reinforces its potential for rapid deployment in situations where data collection is challenging, yet a high degree of accuracy is required.

\subsection{Qualitative Analysis on KIT Actions Dataset}
A comprehensive qualitative analysis was performed to assess the descriptive capabilities of our methods compared to the state-of-the-art methods mentioned earlier. The KIT Actions Dataset was chosen for this analysis due to its size, which made a detailed qualitative evaluation by human raters feasible. Ten evaluators, comprised of bachelor and master students aged between 18 to 29 and well-acquainted with the task, participated in the assessment. Each evaluator was briefed about the objectives of the study to ensure an informed and unbiased rating process.

Participants observed 540 video recordings from the KIT Actions Dataset, along with the corresponding descriptive sentences generated by each method. They were asked to score the outputs based on three criteria: “grammatical correctness,” “semantic correctness,” and “relevance,” with up to 5 points awarded for each category. These criteria were selected to cover the essential aspects of natural language descriptions that contribute to the overall comprehensibility and utility of the generated text.

Table~\ref{tab:qualitative_analysis_kit} summarizes the average scores from the raters for each method across the three evaluation dimensions. Each entry in the table provides a score out of 5 for three categories: Grammatical Correctness, Semantic Correctness, and Relevance.

\begin{table}[ht]
\centering
\caption{Qualitative Analysis of Description Generation on the KIT Bimanual Actions Dataset (Scores Out of 5)}
\label{tab:qualitative_analysis_kit}
\resizebox{\textwidth}{!}{%
\begin{tabular}{|l|c|c|c|}
\hline
Method & Grammatical Correctness  & Semantic Correctness  & Relevance  \\
\hline
Ji et al. \cite{ji2020action}         & 3.2 & 3.1 & 3.0 \\
Krause et al. \cite{krause2017hierarchical} & 3.0 & 3.2 & 3.1 \\
Nabati et al. \cite{nabati2020video} & 3.5 & 3.4 & 3.3 \\
Singh et al. \cite{singh2021efficient} & 3.6 & 3.7 & 3.5 \\
Nguyen et al. \cite{nguyen2018translating} & 3.8 & 3.9 & 3.7 \\
Our Method 1 (Hybrid)                & 4.1 & 4.2 & 3.9 \\
Our Method 2 (End-to-End)            & \textbf{4.3} & \textbf{4.5} & \textbf{4.1} \\
\hline
\end{tabular}
}
\end{table}

The scores were assigned based on human judgment and expert analysis of the output captions. A higher score in each category reflects a closer approximation to human-like performance. For instance, a score of 4.5 in Grammatical Correctness would imply very few grammatical errors, whereas a score of 4.7 in Semantic Correctness would suggest that the generated descriptions closely match the semantic content of the actions being performed.

The results provide insights into the practical effectiveness of each method in generating descriptions that are not only grammatically sound but also semantically accurate and contextually relevant.

Our Methods 1 (Hybrid) and 2 (End-to-End) have been compared against existing approaches to highlight their relative performance. In particular, we observe that Our Method 2 (End-to-End) typically achieves higher scores in the Relevance category, indicating its superior ability to focus on the most crucial elements of the video content. On the other hand, Our Method 1 (Hybrid) shows a strong performance in Semantic Correctness, suggesting that it can generate meaningful captions even with limited data.

These scores further substantiate the adaptability and effectiveness of our proposed methods, providing qualitative insights that align with the quantitative metrics reported earlier. They reflect the balance between producing grammatically sound sentences, capturing the essence of the visual content semantically, and maintaining relevance to the core actions depicted in the videos.

\section{Conclusion}
\label{conclusion}
In this paper, we proposed two methods for producing multiple sentence descriptions of complex manipulation actions, which is a class of actions very common for humans but also prevalent in human-robot interaction. Our central focus had been on the generation of a hierarchically ordered set of different annotations: from detailed and complex to condensed and simple which was achieved by using stacked LSTMs. The problem of generating multiple sentences was studied before e.g.  Senina et al. \cite{rohrbach2014coherent} modeled intra-sentence consistency by considering the probability of occurrence between pairs of objects and actions in consecutive sentences. Zhang et al. \cite{zhang2020relational} proposed a method for multi sentence generation by modeling the relational information between objects and events in the video using a graph-based neural network. More recently, Rohrbach et al. \cite{park2019adversarial} presented an adversarial inference for multi sentence video description.  However, the primary emphasis of existing methods has been on describing the sequence of complex actions executed by one or more subjects and they do not pay attention to the constituent sub-actions that make up each action. Different from this, our work is the first that constructs a hierarchy of action descriptions building on the concept of atomic actions, which are in a video directly recognizable by conventional computer vision methods. This way we can produce descriptive sentences at different levels of granularity. Even transformers may not be the ideal choice to this end as they depend on substantial labeled data and extensive pre-processing procedures. Additionally, they may encounter difficulties in capturing the spatial and temporal relationships in videos. Our joint embedding space approach, on the other hand, employs pre-trained visual and linguistic features to match inputs directly, facilitating efficient generation of multi-sentence descriptions for complex manipulation action videos without extensive pre-processing or large amounts of labeled data.

Quantitative analysis has demonstrated our methods are comparable to the state of the art. Additionally, human raters have confirmed the resulting descriptions as understandable and generally appropriate. In conclusion, we would argue that hierarchical action description should offer additional functionalities allowing users to adopt the descriptive depth to their individual needs.

\subsection{Comparison of Hybrid Statistical with End-to-End Method:\\ Strengths and Limitations}

When selecting an approach for video description generation, it is crucial to consider the strengths and limitations of both the Hybrid Statistical and End-to-End methods. This comparison aims to provide insights for informed decision-making.
Here we summarize the aspects of different methods and discuss their pros and cons. In the Appendix we provide an itemized version of this  to allow for an easier side-by-side comparison.\\

The Hybrid Statistical Method utilizes a combination of handcrafted features, rule-based modeling, and linguistic knowledge to create interpretable descriptions of actions. Its strength lies in providing multi-level descriptions that are not only clear but also based on well-established rules, making it particularly effective for complex actions that follow known patterns. This method, however, is not without its limitations. It relies heavily on manual engineering and linguistic input, which can make it less adaptable to new, diverse actions that are not covered by the predefined rules. While there are strategies to handle unseen data, such adaptations often require manual intervention, which may not be ideal in rapidly changing environments.

On the other hand, the End-to-End Method leverages the power of Convolutional Neural Networks (CNNs) for feature extraction, Gated Recurrent Units (GRUs) for temporal modeling, and Long Short-Term Memory networks (LSTMs) for generating descriptions. This method is, thus, fundamentally data-driven, learning directly from very large datasets, which allows it to generalize across a wide range of tasks and actions. It is scalable and capable of handling unseen data with more ease than the hybrid approach, especially when employing strategies like data augmentation and transfer learning to enhance its adaptability. The end-to-end method is suitable for applications where the availability of large amounts of training data and the need for diverse action handling are paramount, offering automated, data-driven solutions that minimize the need for manual tweaking.

When deciding between the two, the Hybrid Statistical Method is the go-to for scenarios that require highly interpretable, multi-level descriptions for well-defined actions. It's ideal when dealing with complex actions with known patterns and where interpretability is crucial. Conversely, the End-to-End Method is preferred in situations with abundant training data, a need for handling a diversity of actions, and a preference for automated, data-driven solutions. It excels in adaptability, making it capable of handling new actions more effectively. In some cases, combining both methods may offer the best of both worlds, with the hybrid approach tackling complex actions and the end-to-end method providing automation for more common tasks.

\newpage
\setcounter{section}{0}
{\huge APPENDIX}
\section{Spatial Reasoning}
\label{spatial_reasoning}
\subsection{Object modeling by convex hull}
\label{convex_hull}
The smallest bounding (or enclosing) box for a point set $ (S) $ in $ N $ dimensions is the box with the smallest measure (area, volume, or hyper-volume in higher dimensions) within which all the points lie. To provide the required object representation accuracy in this paper, we use the ``3D convex hull'' model which is the smallest convex set that includes the object. We constructed this model according to the ``gift wrapping algorithm'' \cite{sugihara1994robust}.


\subsection{Spatial reasoning using convex hulls}
\label{spatial_relations}
\subsubsection{\textbf{Basic definitions}}
\label{spatial_relations_1}
Here, we propose a model based on space division, including interior, boundary, and exterior points for defining spatial relations. To this end, the spatial topological relations of any two 3D convex hulls can be described in detail by the intersection of their point sets. Suppose $\alpha$ is an object whose point cloud is obtained from a depth camera. We present a formal definition of the boundary,
interior, and exterior of $\alpha$ as follow:\\
\textbf{Definition 1.} The boundary of a point cloud $\alpha$, denoted by $\delta\alpha$, is the convex hull of $\alpha$.\\
\textbf{Definition 2.} The interior of a point cloud $\alpha$, denoted by $\alpha^{0}$, is the interior of $\delta\alpha$,
($\delta\alpha\sqcap\alpha^{0}=\emptyset$).\\
\textbf{Definition 3.} The exterior of a point cloud $\alpha$, denoted by $\alpha^{-}$, is the exterior of $\delta\alpha$. $\alpha^{-}$ does not contain any point of $\alpha$, ($\alpha\sqcap\alpha^{-}=\emptyset$).\\
\textbf{Definition 4.} The closure of a point cloud $\alpha$ is combining the boundary and the interior of the point cloud,
denoted by $\alpha^{+}$, (i.e., $\alpha^{+}=\alpha^{0}\sqcup\delta\alpha$).\\

\begin{figure}[!h]
    \centering
    \includegraphics[width=0.60\textwidth]{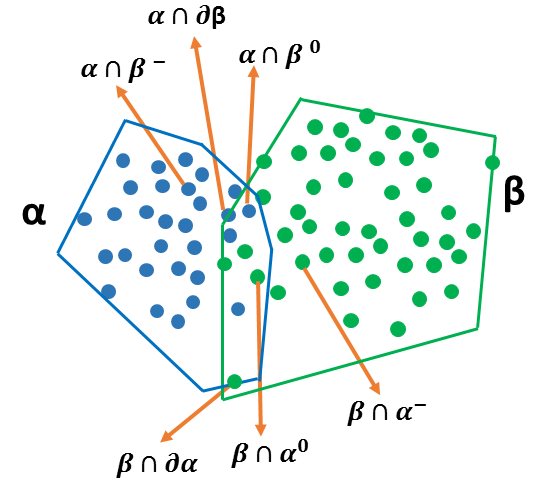}
    \caption{At the intersection of the two convex hulls, the space is divided into 6 separate sections (we show here two 2D convex hulls for simplicity, but this space partitioning is also valid for 3D convex hulls)}
    \label{6IM}
\end{figure}

Accordingly, the 3D-space is completely divided into three parts $\alpha^{0}$, $\delta\alpha$, and $\alpha^{-}$, i.e., ($\alpha^{0}\sqcup\delta\alpha\sqcup\alpha^{-}={\Bbb R}^{3}$). Therefore, the spatial topological relations between two point clouds, namely $\alpha$ and $\beta$, can be represented as the relations between their corresponding interior, boundary, and exterior points and the whole space around is partitioned into six parts (Fig.\ref{6IM}):

\begin{enumerate}
\item the parts of $\alpha$ located in the interior of $\delta\beta$, denoted by $\alpha\sqcap\beta^{0}$;
\item the parts of $\alpha$ located on $\delta\beta$, denoted by $\alpha\sqcap\delta\beta$;
\item the parts of $\alpha$ located at the exterior of $\delta\beta$, denoted by $\alpha\sqcap\beta^{-}$;
\item the parts of $\beta$ located in the interior of $\delta\alpha$, denoted by $\alpha^{0}\sqcap\beta$;
\item the parts of $\beta$ located on $\delta\alpha$, denoted by $\delta\alpha\sqcap\beta$;
\item the parts of $\beta$ located at the exterior of $\delta\alpha$, denoted by $\alpha^{-}\sqcap\beta$;
\end{enumerate}

Thus, the spatial topological relation from $ \alpha $ to $ \beta $ can be represented as a matrix $ Rel(\alpha,\beta):$

\begin{equation}
\label{6IM_mat}
\begin{split}
\begin{pmatrix}
\alpha\sqcap\beta^{0} & \alpha\sqcap\delta\beta & \alpha\sqcap\beta^{-}\\
\alpha^{0}\sqcap\beta & \delta\alpha\sqcap\beta & \alpha^{-}\sqcap\beta
\end{pmatrix}
\end{split}
\end{equation}

Using this matrix, we can define all spatial topological relations from $\alpha$ to $\beta$.

\subsubsection{\textbf{Computations}}
\label{spatial_relations_2}
Spatial relations are abstract relationships between entities in space. A correct understanding of object-wise spatial relations for a given action is essential, for example for a robot to perform an action successfully. Earlier, spatial relations between objects have been divided into ``static'' and ``dynamic'' spatial relations (SSR, DSR) \cite{ziaeetabar2017semantic}. ``Static'' refers to the relations between the location of objects in space, while ``dynamic'' describes the kinetic relationships between objects. A static relation, for example, would be ``Above'' a dynamic one, f.e. ``moving together'': Originally there were 8 static and 6 dynamic relations, which we have now extended (for the complete list as well as a sample of static spatial relations please see Fig. 4 in the main paper).

\begin{figure}[!h]
    \centering
    \includegraphics[width=0.97\textwidth]{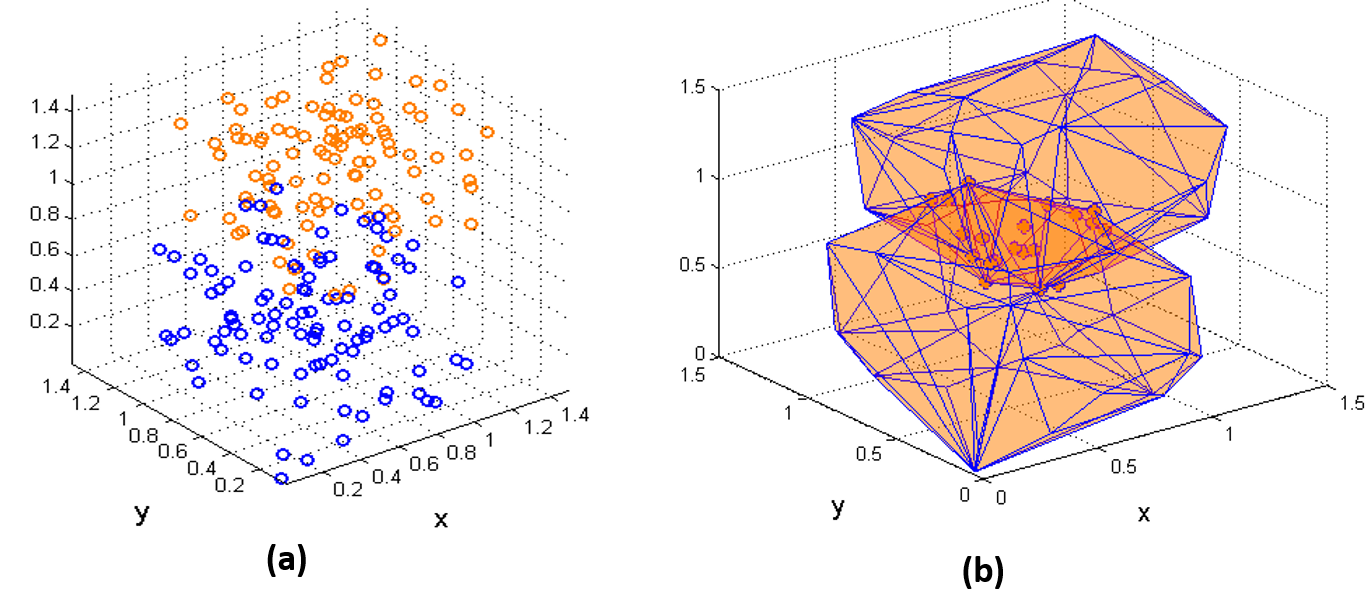}
    \caption{(a) Two object point clouds, (b) the 3D convex hulls of the two point clouds and their touching plane}
    \label{intersection}
\end{figure}

To accurately determine these spatial relations, using convex hulls, we need to compute the matrix $Rel$ for each static relation as explained in equation \ref{6IM_mat}. But before computing $Rel$, we need to assess whether two convex hulls are touching or not touching, because physical contact influences the actual spatial relation. Therefore, we define $Touch(\alpha,\beta)=1$ if the convex hulls touch each other and $Touch(\alpha,\beta)=0$ if not. As our convex hulls are in 3D, their touching area is empty (when they are disjoint) or a plane (when they are touching). A sample of two 3D convex hulls and their touching plane is shown in Fig.\ref{intersection}.

\begin{align}
\begin{split}
\label{touch}
& \textbf{\textit{If} } (\alpha\sqcap\beta^{0}=\emptyset,  \beta\sqcap\alpha^{0}=\emptyset) \wedge (\alpha\sqcap\delta\beta=\neg\emptyset,  \beta\sqcap\delta\alpha=\neg\emptyset)\\ & \Rightarrow  Touch(\alpha,\beta)=1; \\
& \textbf{\textit{If} } (\alpha\sqcap\beta^{0}=\emptyset,  \beta\sqcap\alpha^{0}=\emptyset) \wedge (\alpha\sqcap\delta\beta=\emptyset, \beta\sqcap\delta\alpha=\emptyset)\\ & \Rightarrow  Touch(\alpha,\beta)=0
\end{split}
\end{align}

As illustrated, for computing the $Touch$ we need to know if two sets of points (interior, boundary or exterior) intersect or not. This is done in two steps to speed up computation. First we create the axis-oriented bounding box (AABB) of each object. If a point is located in the convex hull of an object, it must also be located in the AABB of that object. It is easy and fast to evaluate this for all points. If a point is not in the AABB, then this point does not have to be considered any further. For all other points we need to determine whether they are in the convex hull. As mentioned earlier, we obtain the convex hull around of each object by using the gift wrapping algorithm \cite{sugihara1994robust}. This algorithm creates a three-dimensional polyhedron $(\alpha)$, each side of which is a plane. For each of these planes, we check whether the target point lies to the ``left'' of that plane. When performing this, we treat the planes as vectors pointing counter-clockwise around the convex hull. If the target point is to the ``left'' of all of these plane vectors, then it is contained in the polyhedron $(\alpha^{0})$; if the target point is on one of the side planes, it is considered as a polyhedron boundary point $(\delta\alpha)$; otherwise, it lies outside the polyhedron $(\alpha^{-})$. Suppose the equation of plane $N$ is: $aX+bY+cZ+d=0$  and $[ x, y, z]$ are the coordinates of point $P$, then:
\par
\begin{align}
\begin{split}
\label{directions}
& \textbf{\textit{If} }(ax+by+cz+d=0)\Rightarrow p \in N; \\
& \textbf{\textit{If} }(ax+by+cz+d>0)\Rightarrow p \mbox{ is on the right (front) of }N; \\
& \textbf{\textit{If} }(ax+by+cz+d<0)\Rightarrow p \mbox{ is on the left (back) of }N.
\end{split}
\end{align}

This way we have determined $Touch$. We now formulate how to compute each of the static spatial relations defined in the main text Fig. 4:

\begin{align}
\begin{split}
\label{rel1}
& \textbf{\textit{If}} \mbox{ }min(y_{\alpha}) > max(y_{\beta});\\
& \textbf{\textit{If}} \sim(max(x_{\alpha}) < min(x_{\beta}) \parallel min(x_{\alpha}) < max(x_{\beta}));\\
& \textbf{\textit{If}} \sim(max(z_{\alpha}) < min(z_{\beta}) \parallel min(z_{\alpha}) < max(z_{\beta}));\\
& \Rightarrow SSR(\alpha,\beta)=\textbf{\textit{Ab}}, \mbox{ }SSR(\beta,\alpha)=\textbf{\textit{Be}}
\end{split}
\end{align}

\begin{align}
\begin{split}
\label{rel2}
&\textbf{\textit{If}} \mbox{ }SSR(\alpha,\beta)= Ab \mbox{ } \wedge \mbox{ } Touch(\alpha,\beta)=1;\\
&\Rightarrow SSR(\alpha,\beta)= \textbf{\textit{To (Bo)}}
\end{split}
\end{align}

Since the maximum and minimum coordinates (x, y, and z) of the convex hulls are the same as the maximum and minimum coordinates of the objects, it is easy to check conditions such as those in equation \ref{rel1}.

``Left'' (L) and ``Right'' (R) relations as well as ``Front'' (F) and ``Back'' (Ba) relations between two convex hulls are also defined in a manner similar to the method of equation \ref{rel1}. These four relations are merged into ``Around'' (Ar) which is converted to ``Around with touch'' (ArT) if the convex hulls touch each other according to equation \ref{touch}.\\

Moreover, to compute  ``Cross'' (Cr), ``Within'' (Wi), ``Partial within'' (Pwi), ``Contain'' (Co) and ``Partial contain'' (Pco) between two objects, first they are modeled in term of convex hulls and then their $ Rel $ matrix (equation \ref{6IM_mat}) is computed.

\textbf{Cross:}
\begin{equation}
\label{cross}
\begin{split}
SSR (\alpha,\beta)= \textbf{\textit{Cr}} \Leftrightarrow Rel(\alpha, \beta)=
\begin{pmatrix}
\neg\emptyset & \ast \mbox{ }  \mbox{ }  \ast\\
\neg\emptyset & \ast  \mbox{ }  \mbox{ } \ast
\end{pmatrix}
\end{split}
\end{equation}

\textbf{Within:}
\begin{equation}
\label{within}
\begin{split}
SSR (\alpha,\beta)= \textbf{\textit{Wi}} \Leftrightarrow Rel(\alpha, \beta)=
\begin{pmatrix}
\neg\emptyset & \ast \mbox{ }  \mbox{ }  \emptyset\\
\emptyset & \ast  \mbox{ }  \mbox{ } \ast
\end{pmatrix}
\end{split}
\end{equation}

\textbf{Partial within:}
\begin{equation}
\label{partial_within}
\begin{split}
SSR (\alpha,\beta)= \textbf{\textit{Pwi}} \Leftrightarrow Rel(\alpha, \beta)=
\begin{pmatrix}
\neg\emptyset & \ast \mbox{ }  \mbox{ }  \neg\emptyset\\
\emptyset & \ast  \mbox{ }  \mbox{ } \ast
\end{pmatrix}
\end{split}
\end{equation}

\textbf{Contain:}
\begin{equation}
\label{contain}
\begin{split}
SSR (\alpha,\beta)= \textbf{\textit{Co}} \Leftrightarrow Rel(\alpha, \beta)=
\begin{pmatrix}
\emptyset & \ast \mbox{ }  \mbox{ }  \ast\\
\neg\emptyset & \ast  \mbox{ }  \mbox{ } \emptyset
\end{pmatrix}
\end{split}
\end{equation}

\textbf{Partial contain:}
\begin{equation}
\label{partial_contain}
\begin{split}
SSR (\alpha,\beta)= \textbf{\textit{Pco} }\Leftrightarrow Rel(\alpha, \beta)=
\begin{pmatrix}
\emptyset & \ast \mbox{ }  \mbox{ }  \ast\\
\neg\emptyset & \ast  \mbox{ }  \mbox{ } \neg\emptyset
\end{pmatrix}
\end{split}
\end{equation}

In all of the above equations, ``$\emptyset$'' and ``$\neg\emptyset$'' represent the empty and non-empty sets, respectively while ``$\ast$'' sign indicates an arbitrary value.

\section{Decomposition of Complex Manipulation Actions into Atomic Actions}
\label{decompose}
In section 5.1 of the main paper, we introduced concept of atomic actions as a cornerstone of complex manipulation actions. Each atomic action is represented using a quintuple including information about its subject, action type, object, spatial relation between subject and object as well as the place of occurrence. Action types have only 4 different, simple options which are ``Touching (T)'', ``Untouching (U)'', ``Moving together (Mt)'' and ``Fixed moving together (Fmt)''. Real actions like ``cut'', thus, consist of a chain of atomic actions. Hence, we need a framework that is able to decompose complex ``real'' manipulation actions into their constituent atomic actions which then provides an efficient mechanism for their representation and recognition. We use a Context-Free Grammar (CFG) for this purpose.

A CFG is defined as $G = \langle S,N,T,P \rangle$, where $S$ is the start point, $N$ is list of non-terminals and $T$ contains list of terminals. Moreover, $P$ includes list of production rules. In a CFG, every production rule is of the form: $A\rightarrow\alpha$, where $A$ is a single non-terminal symbol, and $\alpha$ is a string of terminals and/or non-terminals. A formal grammar is considered ``context free'' when its production rules can be applied regardless of the context of a non-terminal. No matter which symbols surround it, the single non-terminal on the left-hand side can always be replaced by the right-hand side.\\

We can use this structure to map it to manipulation actions, because, similarly, here every complex action is built from smaller blocks, where each block is an atomic action or a sequence of those. Furthermore, a complex action also has a basic recursive structure, and can be decomposed into simpler actions. Hence, intuitively, the process of observing and interpreting manipulation actions, is syntactically  similar to natural language understanding \cite{yang2014cognitive}. Therefore, we borrow the structure of a Context-Free Grammar for the purpose of parsing complex manipulation actions to their constituent components, which also leads to an efficient way of action representation. Within this scope, ``hand(s)'', list of ``atomic action types'', ``objects'', ``spatial relations'' and ``places'' are defined as terminals.  Moreover, ``$S_p$'' or ``sentence phrase'', ``$O_p$'' or ``object phrase'', ``$A_p$'' or ``action phrase'' and ``$SR_p$'' or ``spatial relation phrase'' are our non-terminals, which, with their reversible properties, create proper representations of the possible sequences of atomic actions. Another non-terminal is ``Sub'', which indicates the performer of the action, which can be a hand or a combination of the hand and the object being touched by the hand for a purpose (e.g. a tool). The latter is called merged entity (Me).``S'' is the start point from which the construction of all action sequences begins.
\begin{align}
\label{mygrammar}
\begin{split}
& S \rightarrow S_p | S_p.S_p  \\
& S_p \rightarrow Sub.A_p | S_p.Sub.A_p \\
& Sub \rightarrow Hand | Me \\
& A_P \rightarrow A.O_P|A_P.O_P\\
& Me \rightarrow Hand.O\\
& O_P \rightarrow O.{SR}_P\\
& {SR}_P \rightarrow SR.P\\
& Hand \rightarrow Hand_L | Hand_R | Me\\
& A \rightarrow A_1 | A_2 | ... | A_{183^*}\\
& O \rightarrow O_1 | O_2 |... (list\,of\,objects)\\
& P \rightarrow Ground|Air| O_1 | O_2 |... (list\,of\,objects)\\
& SR \rightarrow Ab | Be | To | Bo | Ar | ArT | Cr | Wi | Pwi | Co | Pco\\
\end{split}
\end{align}

\begin{small}
*: The total number of valid permutations between components of an atomic action is 183. Therefore, we will have 183 different atomic actions that all have been stored in a repository called \textbf{\textit{library of action mappings}}.\\
\end{small}

\begin{figure}[!h]
    \centering
    \includegraphics[width=0.98\textwidth]{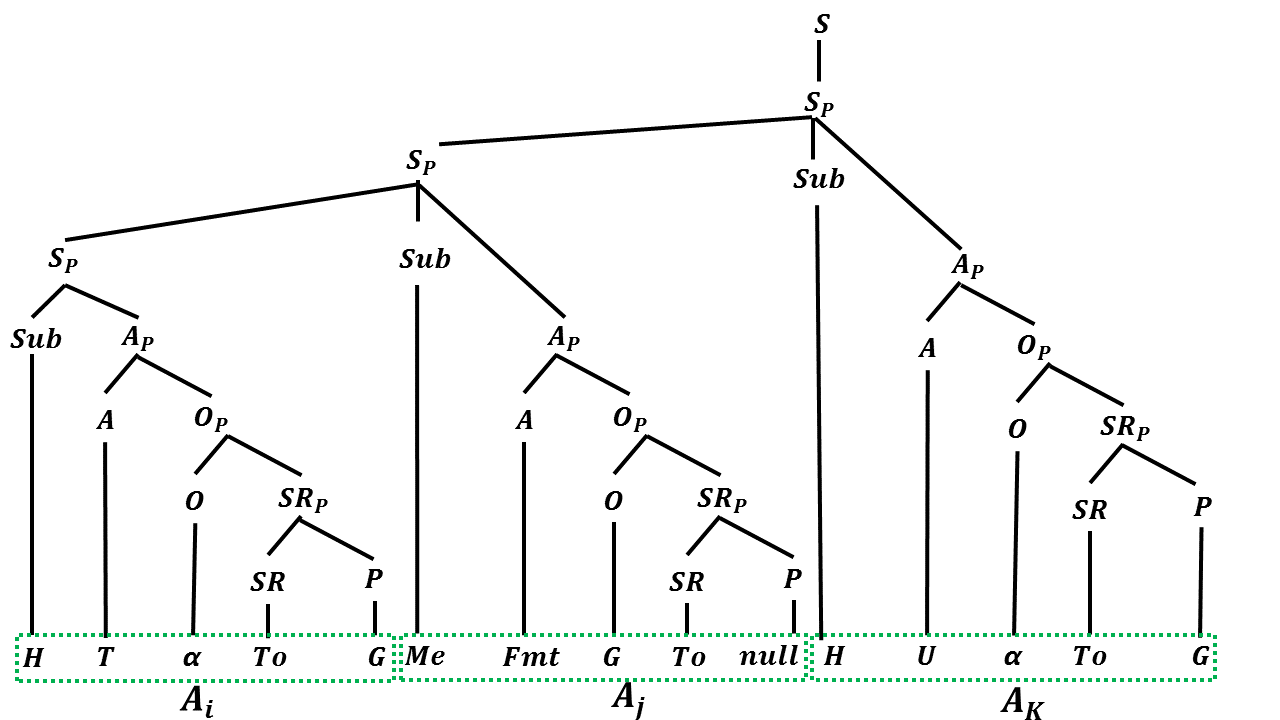}
   \caption{Decomposition of a \textit{push} action into its components according to the CFG rules (equation \ref{mygrammar})}
   \label{push}
\end{figure}

Fig. \ref{push} contains a pars tree showing the decomposition of a manipulation - here \textit{push} -  to its constituent atomic actions using the above defined CFG rules.

\section{Results}
Here, we first report quantitative results from the implementation of the spatial reasoning algorithm based on the convex-hull modeling. Then we will review complex manipulation action recognition results obtained using the proposed context-free grammar.

\subsection{Spatial reasoning}
In order to evaluate the improvement that is caused by applying convex-hull modeling compared to AABB (Axis-Aligned bounding box) which has been used in previous works \cite{ziaeetabar2017semantic} for detecting correct spatial relations, we used the ``KIT Bimanual Actions Dataset'' , \cite{dreher2020learning}. This dataset includes $540$ recordings, where each contains various objects with different spatial relations. These relations change during the progress of an action along the video frames. For example in a cutting scenario, in the beginning, the knife is located  ``around'' (near) a fruit, then it touches its top, then it goes inside the fruit, etc. 
In this way, we consider every ten frames in each scenario and obtained the spatial relations between each pair of objects (approximately five pairs in each video clip) using both convex-hulls and AABBs. In this dataset, humans perform their movements at a normal speed, hence, this procedure is fast enough to recover all changes in spatial relations.
The results indicated that among of $ \textbf{110500} $ cases, and compared to the ground truth provided in the dataset for the previously defined SSRs (\cite{ziaeetabar2017semantic}) and the additional ground truth provided by us for the newly defined SSRs using convex-hulls, our object models including convex-hulls and AABBs act successfully in $\textbf{100665\:(91.1\%)}$ and $\textbf{83427\,(75.7\%)}$ cases, respectively. In addition to this, as a non negligible improvement, now also the relations such as \textit{Cr}, \textit{Wi}, \textit{Pwi}, \textit{Co}, \textit{Pco} can be distinguished by convex-hull modeling in contrary to the AABB-based approach. The resulting accuracy is important with respect to two aspects, it is required for \textbf{(1)} the correct production of the semantic representations (SRes) as the basis of our hybrid statistical method, and \textbf{(2)} for the generation of more precise and human-like video descriptive sentences.

\subsection{Complex manipulation actions recognition}
The ``KIT Bimanual Actions Dataset''\cite{dreher2020learning} contains 540 recordings where 6 subjects performed 9 tasks (5 in a kitchen context, 4 in a workshop context) with 10 repetitions. In this dataset there are $221000$ RGB-D image frames with $640\,px*480\,px$ image resolution, which create videos with a total length of $2$ hours and $18$ minutes. The tasks are a combination of 14 sub-tasks where each of the sub-tasks can be an atomic action (e.g. lift) or a mixture of several atomic actions (e.g. cut or stir). Actions are fully labeled for both hands separately. 
Table \ref{res1} shows the list of manipulations in the dataset as well as the accuracy of our CFG based approach in their classification. The grammar used (equation \ref{mygrammar}) decomposes complex actions of each hand (left or right) into their constituent atomic actions and then based on the \textbf{(1)} sequence of identified ``atomic actions'' and \textbf{(2)} referring to the \textit{library of action mappings}, the manipulation type is recognized. 

\begin{table}
    \caption{Classification accuracy according to the CFG based approach on the ``KIT Bianual Action Dataset, \cite{dreher2020learning}''}
         \begin{tabular}{lc}
\hline
\textbf{Manipulation actions} & \multicolumn{1}{l}{\textbf{Classification accuracy}} \\ \hline
Idle                          & 100\%                                                \\ \hline
Approach                      & 100\%                                                \\ \hline
Retreat                       & 100\%                                                \\ \hline
Lift                          & 100\%                                                \\ \hline
Place                         & 97\%                                                 \\ \hline
Hold                          & 100\%                                                \\ \hline
Stir                          & 94\%                                                 \\ \hline
Pour                          & 92\%                                                 \\ \hline
Cut                           & 90\%                                                 \\ \hline
Drink                         & 89\%                                                 \\ \hline
Wipe                          & 91\%                                                 \\ \hline
Hammer                        & 94\%                                                 \\ \hline
Saw                           & 90\%                                                 \\ \hline
Screw                         & 88\%                                                 \\ \hline
\textbf{Average}              & \textbf{95\%}                                        \\ \hline
\end{tabular}
       \label{res1}
    \end{table}

Classifying manipulations at an acceptable percentage $\textbf{(~95\%)}$ promises to produce more accurate video descriptions with less errors also outside the proposed framework.

\section{Itemized comparison of the different methods}

\subsection{Hybrid Statistical Method}

\textbf{Approach:} The Hybrid Statistical Method combines handcrafted features, rule-based modeling, and linguistic knowledge.\\

\noindent
\textbf{Strengths:}
\begin{itemize}
    \item \textbf{Interpretable, Rule-Based Descriptions:} This method provides interpretable descriptions based on predefined rules.
    \item \textbf{Multi-Level Descriptions:} It supports multi-level descriptions, from atomic actions to comprehensive narratives.
    \item \textbf{Effective for Known Patterns:} Particularly effective for complex actions with well-understood patterns.
    \item \textbf{Handling Unseen Data:} Strategies can be employed to adapt to new, unseen actions, although manual intervention may be required.
\end{itemize}

\noindent
\textbf{Limitations:}
\begin{itemize}
    \item \textbf{Manual Engineering:} It relies on manual feature engineering and linguistic knowledge.
    \item \textbf{Limited Adaptability:} May struggle with diverse, unseen actions not covered by predefined rules.
\end{itemize}

\subsection{End-to-End Method}

\textbf{Approach:} The End-to-End Method employs Convolutional Neural Networks (CNNs) for feature extraction, Gated Recurrent Units (GRUs) for temporal modeling, and Long Short-Term Memory networks (LSTMs) for description generation.\\

\noindent
\textbf{Strengths:}
\begin{itemize}
    \item \textbf{Data-Driven Learning:} It is data-driven, learning directly from data without the need for manual engineering.
    \item \textbf{Generalizability:} Generalizes well to diverse actions and descriptions, making it suitable for a wide range of tasks.
    \item \textbf{Scalability:} Scalable and effective when handling large datasets.
    \item \textbf{Handling Unseen Data:} Capable of adapting to new actions or scenarios, provided they share similarities with actions seen during training. Strategies like data augmentation and transfer learning can enhance this capacity.
\end{itemize}

\subsection{Choosing the Right Method}

The choice between the Hybrid Statistical and End-to-End methods should align with specific application requirements:\\

\noindent
\textbf{Use Hybrid Statistical Method when:}
\begin{itemize}
    \item Well-defined actions and multi-level descriptions are needed.
    \item Interpretability is a critical requirement.
    \item Dealing with complex actions with known patterns.
    \item Handling Unseen Data: It can adapt to some extent with manual intervention.
\end{itemize}

\textbf{Use End-to-End Method when:}
\begin{itemize}
    \item Abundant training data is available.
    \item Diverse actions need to be handled effectively.
    \item Automated, data-driven solutions are preferred.
    \item Handling Unseen Data: It offers adaptability and can leverage strategies to handle new actions.
\end{itemize}

In some scenarios, a combination of both methods might be suitable, where the hybrid method handles complex actions, and the end-to-end method automates descriptions for more common or numerous actions.

\newpage
{
\bibliographystyle{ieee_fullname}
\bibliography{egbib}
}

\end{document}